\documentclass[conference]{IEEEtran}
\usepackage{amsmath} 
\usepackage{amssymb}  
\usepackage{boldline}
\usepackage{array,multirow}
\usepackage{dblfloatfix} 
\usepackage{hyperref}
\usepackage{float}
\usepackage{color}
\usepackage{xfrac}
\usepackage{graphicx}
\graphicspath{ {images/} }
\definecolor{light-gray}{gray}{0.95}
\newcommand{\code}[1]{\colorbox{light-gray}{\texttt{#1}}}
\DeclareMathOperator*{\argmax}{arg\,max}

\usepackage{makecell}
\usepackage{bbm}
\usepackage[table,xcdraw]{xcolor}
\usepackage[flushleft]{threeparttable}
\usepackage[utf8]{inputenc}
\usepackage{capt-of}
\usepackage{csquotes}
\usepackage{caption}
\usepackage{algorithm}
\usepackage[noend]{algpseudocode}
\usepackage{multirow}
\usepackage[table,xcdraw]{xcolor}
\usepackage[normalem]{ulem}
\useunder{\uline}{\ul}{}
\hypersetup{
    colorlinks=true,
    linkcolor=blue,
    filecolor=magenta,      
    urlcolor=cyan,
}

\begin{document}
\pagenumbering{arabic}
\title{Enhancing Image Captioning with Neural Models}
\author{\IEEEauthorblockN{1\ Pooja Bhatnagar}
\IEEEauthorblockA{impooja37@gmail.com}
\and
\IEEEauthorblockN{2\ Sai Mrunaal}
\IEEEauthorblockA{saimrunaal@gmail.com}
\and
\IEEEauthorblockN{3\ Sachin Kamnure}
\IEEEauthorblockA{s.kamnure@gmail.com}
}

\nocite{vinyals}
\nocite{tanti1}
\nocite{tanti2}
\nocite{cho}
\nocite{kumar}
\nocite{xu}
\nocite{8k}
\nocite{30k}
\nocite{coco}
\nocite{wu}
\nocite{vgg}
\nocite{keras}
\nocite{imagenet}
\nocite{zipf}
\nocite{word2vec}
\nocite{bleu}
\nocite{rouge}
\nocite{cider}

\maketitle
\thispagestyle{empty}
\pagestyle{empty}

\begin{abstract}
This research explores the realm of neural image captioning using deep learning models. The study investigates the performance of different neural architecture configurations, focusing on the "inject" architecture, and proposes a novel quality metric for evaluating caption generation. Through extensive experimentation and analysis, this work sheds light on the challenges and opportunities in image captioning, providing insights into model behavior and overfitting. The results reveal that while the "merge" models exhibit a larger vocabulary and higher ROUGE scores, the "inject" architecture generates relevant and concise image captions. The study also highlights the importance of refining training data and optimizing hyperparameters for improved model performance. This research contributes to the growing body of knowledge in neural image captioning and encourages further exploration in the field, emphasizing the democratization of artificial intelligence.
\end{abstract}

\section{INTRODUCTION}

The goal of image captioning systems is to generate relevant descriptions of images automatically. This task is substantially more complex than more well-studied computer vision tasks - such as image classification and object detection - as an adequate image captioning system needs to not only identify the objects in an image and how they relate to one another but also incorporate this perceptual information into a language model which can generate descriptions in coherent natural language. 

Image captioning systems have many industrial applications, such as helping visually impaired users understand a website's content, automatic subtitle generation, or identifying the semantic roles of objects within an image. Image captioning is also of theoretic value, as solving the task would represent a step towards complete scene understanding - where a machine can ``see" and interpret perceptual information as a human does - a long-standing dream of computer vision and artificial intelligence practitioners. 

The recent success of neural network approaches to statistical machine translation has inspired researchers to develop end-to-end neural models for image captioning that are fully trainable by backpropagation. Aided by large image classification datasets (Russakovsky et al.; 2015) and powerful networks trained to solve various computer vision tasks, it is possible to encode the contents of an image into a compact vector representation. This representation can then be combined with linguistic features to generate (in other words, be \textit{decoded} to) novel captions. Thus, image captioning systems broadly fall under the Encoder-Decoder neural network framework. 

Previous academic work of such systems typically utilizes datasets specially crafted for the image classification task, such as Flickr-8k (Rashtchian et al.; 2010) and Flickr-30k (Young et al., 2014). In contrast, our system is trained on a \href{https://www.yelp.com/dataset}{real-world image dataset} collected by Yelp as part of the $11^{\text{th}}$ round of the Yelp Dataset Challenge. We discuss the challenges of working with datasets not specialized for image captioning tasks and analyze the root causes for some unexpected behaviors that stem from challenges. 

Our system often produces reasonable captions and clearly considers both the input images' content and the coherence of the output caption. Perhaps more importantly, by describing our methodology, detailing the experiments we ran, and performing both quantitative and qualitative analysis of our results, we reveal some of the important considerations one should be aware of when building an image captioning system on a real-life dataset, including:
\begin{itemize}
\item Different approaches to incorporating linguistic and perceptual features and how to structure a dataset so that it can be used to train a neural image captioning system.
\item How to formalize the tradeoff between caption relevance and conciseness during the generation process.
\item Difficulties that arise from working with data collected organically - i.e., without prompting subjects to generate examples with detailed instructions.
\item Common metrics for automatically evaluating an image captioning system and its motivations.
\item Unexpected behaviors result from working with a sparse corpus with a large vocabulary - a common challenge amongst many natural language datasets. 
\end{itemize}

\section{RELATED WORK}

In recent years, the deep learning community has succeeded in core computer vision tasks, such as image classification, object identification, and image segmentation (Sinha et al. 2018). These successes, coupled with advancements in neural machine translation and language modeling technologies, have enabled researchers to develop end-to-end neural network models that can be used for image captioning systems. 

Vinyals et al (2014). describe such a system, which utilizes the power of Convolutional Neural Networks (CNN) for extracting complex visual features from images and Recurrent Neural Networks (RNN) for language modeling tasks. Inspired by the encoder-decoder framework for machine translation, the authors use a deep CNN to encode images to fixed-length vector representation and feed this representation to an RNN to generate a caption (further details in section III).

In their 2017 papers, Tanti et al. describe a set of architectures that fall under the general encoder-decoder framework. The primary difference between these architectures concerns how to feed the image features to the RNN cells, if at all. Tanti et al. rigorously test the efficacy of image captioning models with varying architectures on canonical image captioning datasets in an attempt to deduce favorable architectures for image captioning models and to form a conjecture as to what broad tasks RNNs are best suited for.

Xu et al. (2016) achieve state-of-the-art performance on standard datasets by incorporating an attention mechanism. The authors argue that attention is useful because it enables a network to focus on the important features of an image and ignore noisy features common in cluttered images. Attention also provides additional interpretability to an image captioning model, as one can visualize what section of an image the attention is focused on at each time step. 

Previous work focuses on the use of specialized datasets that were collected for image captioning and object segmentation tasks, such as Flickr-8k (Rashtchian et al.; 2010), Flickr-30k (Young et al.; 2014), and MS-COCO (Lin et al.; 2014). These datasets were collected by giving human annotators detailed instructions on annotating each image, resulting in relatively clean datasets. This work uses a much more complex, real-world dataset collected by Yelp. As Yelp users are given no instructions on annotating each image before uploading it to the site, this dataset makes it much more difficult to learn a compelling image captioning network. In this work, we discuss difficulties and undesirable behaviors when training an image captioning network on such data. We also experiment with several architectures proposed by Tanti et al. to study which architectures perform best under such circumstances.

\section{THEORETICAL BACKGROUND}

In this section, we discuss several abstract concepts regarding the problem of Neural Image Captioning (NIC) which motivates many of the design choices and experiments that follow in later sections of this report. First, we discuss fitting the image captioning problem into the probabilistic framework. Then, we briefly describe encoder-decoder models and how the models we built are inspired by using the encoder-decoder framework in neural machine translation. Finally, we highlight an important design choice made when specifying the architecture of an NIC model and the conceptual implication of this choice. 

\subsection{Training as Maximizing Caption Likelihood}

Although much of the previous work on image and video captioning involves stitching together several independent systems, researchers are achieving increasing success in building single neural networks for image captioning that are fully trainable by backpropagation. Learning the parameters of such models has a natural probabilistic interpretation; the objective is to directly maximize (with respect to the model parameters, $\theta$) the joint probability of the predicted captions given input images, $I$. For a dataset of $N$ (image, caption) pairs, this can be modeled as\footnote{Under the assumption that the correct captions of the images are mutually independent.}:

\begin{equation}
\theta_{\text{MLE}} = \argmax_{\theta}{\sum_{i = 1}^N{\log P(S^{(i)}|I^{(i))}; \theta)}}
\end{equation} 

Where $(S^{(i)}, I^{(i)})$ is the $i^{\text{th}}$ (image,caption) pair, and each caption $S^{(i)}$ is a sequence of tokens $(S^{(i)}_1, S^{(i)}_1, ... S^{(i)}_k)$ for a fixed maximum sequence length, $k$. 

We may expand (1) by the chain rule of probability:

\begin{equation}
\theta_{\text{MLE}} = \argmax_{\theta}{\sum_{i = 1}^N{\sum_{t = 1}^k{\log P(S_t^{(i)}|S_{t - 1}^{(i)} ... S_{1}^{(i)}; \theta)}}}
\end{equation} 

Equation (2) is the inspiration for the training schema we used to learn the weights, to be discussed further in sections 5.C and 6.A.

\subsection{Encoder-Decoder Framework for Image Captioning}

\begin{figure*}[h]
\centering
\begin{minipage}{.5\textwidth}
  \centering
  \includegraphics[width=.65\linewidth]{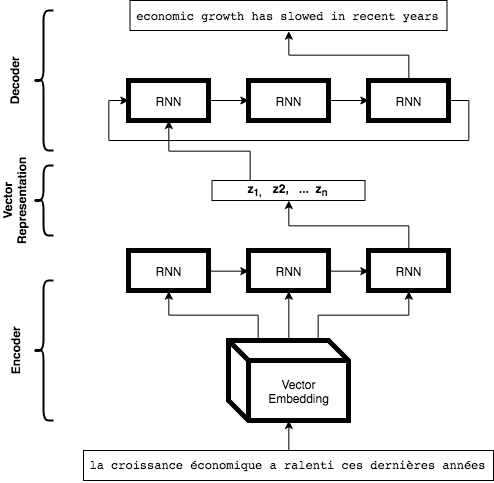}
  \label{fig:test1}
\end{minipage}%
\begin{minipage}{.5\textwidth}
  \centering
  \includegraphics[width=.65\linewidth]{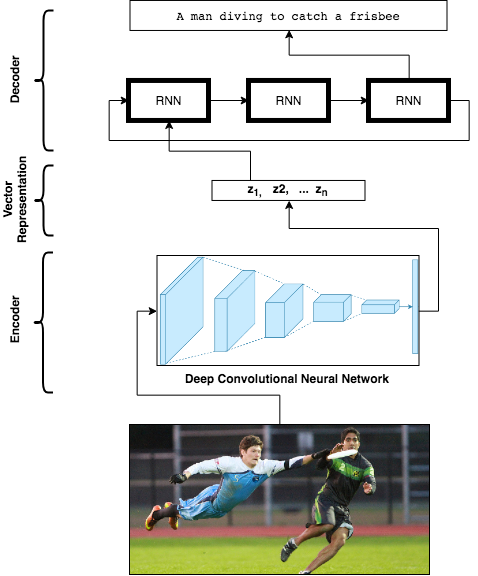}
  \label{fig:test2}
\end{minipage}
\caption{The encoder-decoder framework in neural machine translation models (left) and image captioning models (right). In machine translation applications, an input sequence is encoded by an RNN to a vector representation before decoded to an output sequence. Analogously, a neural image captioning system uses a CNN to encode an input image before generating an output sequence.}
\end{figure*}

Traditional sequence prediction tasks - such as time series forecasting and language modeling - involve using a varying length sequence of inputs (whose ordering usually has a temporal interpretation), and predicting a single output (usually at the next time step). Neural networks are well equipped to solve this type of ``many-to-one" problem, as sequences of inputs may be coerced to a fixed number of time steps so that the network does not need to account for varying length inputs or outputs.

Another sequence prediction task class is those with varying length sequences as inputs and outputs. These so-called ``many-to-many" or ``sequence-to-sequence" problems are more difficult than many-to-one problems, as the network must learn to produce predictions of varying lengths.

An important example of a sequence-to-sequence problem is that of statistical machine translation (SMT). SMT models attempt to take in a sequence of words in one language and output a sequence of words in another while preserving the meaning of the input sequence and the coherence of the output sequence.

An approach that has proven effective for machine translation is called the ``Encoder-Decoder" architecture. This architecture comprises two parts: an ``encoder," which takes in a varying length input sequence and encodes it into a fixed-length vector representation. A ``decoder" model decodes this vector representation into a varying-length sequence prediction. 

Cho et al. (2014) use this framework to build an SMT system that translates sequences from French to English. They use LSTM recurrent neural networks to encode input sequences and LSTMs to decode the encoded representations to form the predictions. Cho et al. show that the encoded representations of input sequences preserve syntactic and semantic structure - thus, this representation is often referred to as a ``sequence embedding." Variants of encoder-decoder models have become the state of the art in machine translation; in fact, Google has adopted this approach in their Google Translate Service (Wu et al.; 2016).

Modern neural image captioning (NIC) systems, including those described in this paper, adopt this approach for the problem of image captioning. Image captioning is similar to machine translation in that the output is a varying-length sequence of natural language but differs in that the input is an image rather than a sequence. Thus, instead of using RNN models to encode the input, NIC systems typically use CNN models to form a vector representation from an image. Recurrent neural networks then use this representation to form an output sequence. Thus, NIC systems fall under the Encoder-Decoder framework, with CNNs as encoders and RNNs as decoders [figure 1].

The use of CNNs as the success of CNNs in core computer vision tasks, such as image classification and segmentation, justifies encoders. CNNs succeed in these tasks because they can effectively extract meaningful and complex features from images. Thus, just as LSTMs encode the semantic and syntactic of an input sequence in SMT applications, it is believed that CNNs encode meaningful properties of an input image in NIC systems.

\subsection{Merge and Inject Models}

In a NIC system, CNN image features are combined with previously predicted tokens to predict the next token in a sequence. There are many ways to condition this sequence prediction on the image features, each of which assigns a different role to the RNN component of a NIC system. 

In their paper \emph{What is the Role of Recurrent Neural Networks (RNNs) in an Image Caption Generator?}, 
Tanti et al. describe two important classes of NIC models that differ in their method of incorporating image features. 

The first, referred to as ``Condition-by-Inject" (or ``Inject" models for short), incorporates the CNN image features directly into the RNN component. Using both the image features and the previously predicted tokens to predict the next token of the sequence. This is usually done by conditioning the internal state of the RNN cells with the image representation or treating the image representation as the first `word' in the input sequence. The second method, ``Condition-by-Merge" (or ``Merge" models for short), never incorporates the image features into the RNN model. Instead, the RNN is used to encode the linguistic features from previously predicted tokens, independent of the perceptual encoding formed by the CNN. Then, the linguistic and perceptual encodings are merged into a single representation, from which the output prediction is made [figure 2]. 

\begin{figure*}[h]
\centering
\begin{minipage}{.5\textwidth}
  \centering
  \includegraphics[width=.85\linewidth]{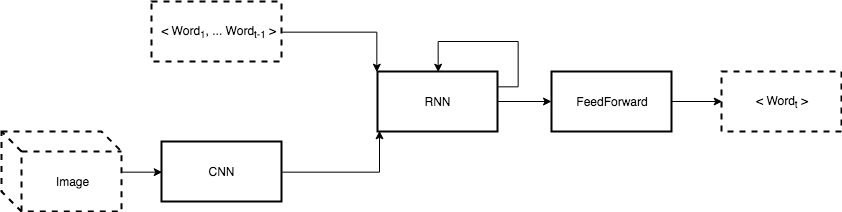}
  \label{fig:test1}
\end{minipage}%
\begin{minipage}{.5\textwidth}
  \centering
  \includegraphics[width=.85\linewidth]{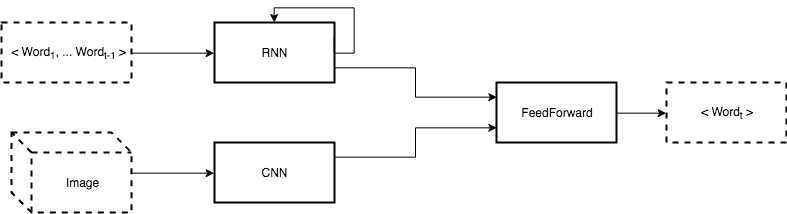}
  \label{fig:test2}
\end{minipage}
\caption{A bare-bones diagram of the Inject architecture (left) and Merge architecture (right). The former conditions the RNN with both the image and linguistic features and uses the RNN to generate new tokens. The uses the RNN to encode the previously predicted tokens into a vector representation rather than to generate new tokens.}
\end{figure*}

The Inject and Merge architectures assign different conceptual roles to the RNN in a NIC system. In the Inject model, perceptual and linguistic features are available to the RNN, responsible for generating the next predicted token. Thus, the RNN acts as a ``generator" in the Inject architecture. In contrast, the RNN has no access to the image features in the Merge architecture. Instead, the RNN encodes an input sequence into a fixed-length vector representation, and a later fully connected layer generates new tokens. Thus, the RNN can be considered an encoder rather than a generator in the Merge architectures. 

In the experiments described in section 6, we compare models using the Inject and Merge architectures. The idea that the choice of architecture dictates the `role' of the RNN in a NIC adds a new perspective to these experiments. The relative efficacy of one architecture over another not only shows us which architecture is more appropriate for this task but provides insights into what settings RNNs are best suited for in general.  

\section{The Yelp Dataset}

Yelp maintains \href{https://www.yelp.com/dataset}{a public dataset} for research and education. Part of this dataset is a collection of over 200,000 images uploaded by users, 100,807 of which contain captions written by users upon uploading these images. This subset of images is what we used our train our image captioning system. Images are annotated with one of the categories \textit{``food", ``inside", ``outside", ``menu"} or \textit{``drink"}.

Compared to standard benchmark datasets for image captioning, such as Flickr-8k and Flickr-30k, the Yelp dataset makes for a much more difficult task in training an effective NIC model.

Standard datasets typically have 4-8 training captions per image, which helps disambiguate which words are related to an image's content and which words result from more noisy phenomena, such as an author's writing style. 

More importantly, datasets such as Flickr-8k and Flickr-30k were collected by giving human subjects detailed instructions on how to form a training caption [figure 3]. In contrast, Yelp users are given no guidance on what photos to upload to the site nor how to write captions for these photos. The result is a dataset with much larger caption style, content, and relevance variability. More severely, since the users are not given a common set of instructions on constructing training captions, modeling the image-caption relationship is not well defined. Figure 4 shows four images that exemplify why training a NIC on the Yelp image dataset is difficult. 

\begin{figure}[]
\centering
\includegraphics[width=1\linewidth]{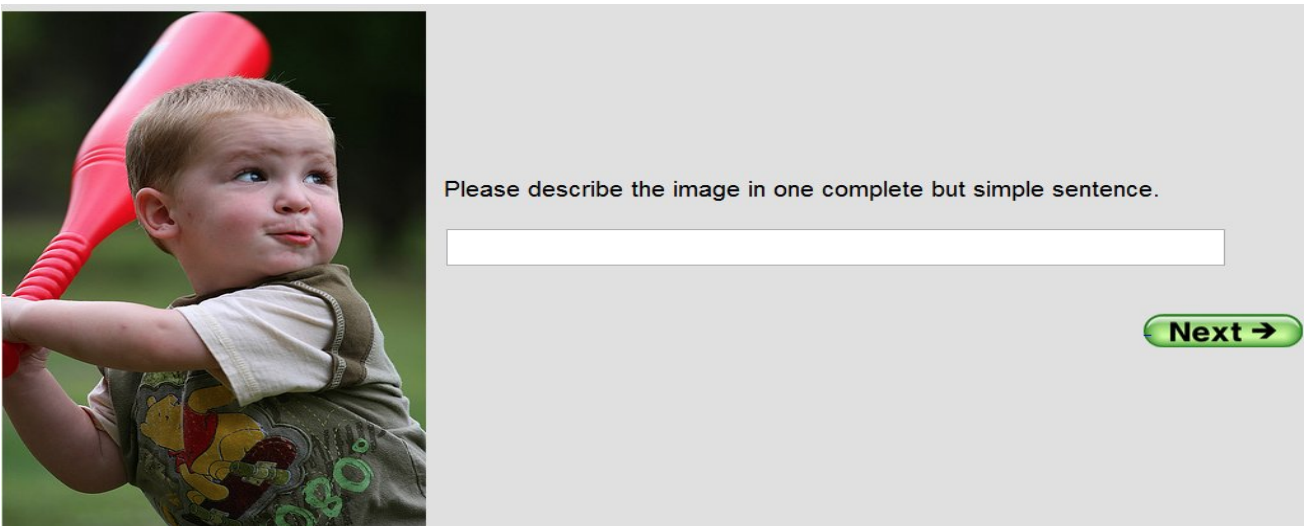}
\caption{Instructions presented to Amazon Turkers in the collection of the Flickr-8k dataset (Rashtchian et al., 2010). Image extracted directly from [7].}
\label{fig:test1}
\end{figure}

\begin{figure}[]
\centering
\includegraphics[width=1\linewidth]{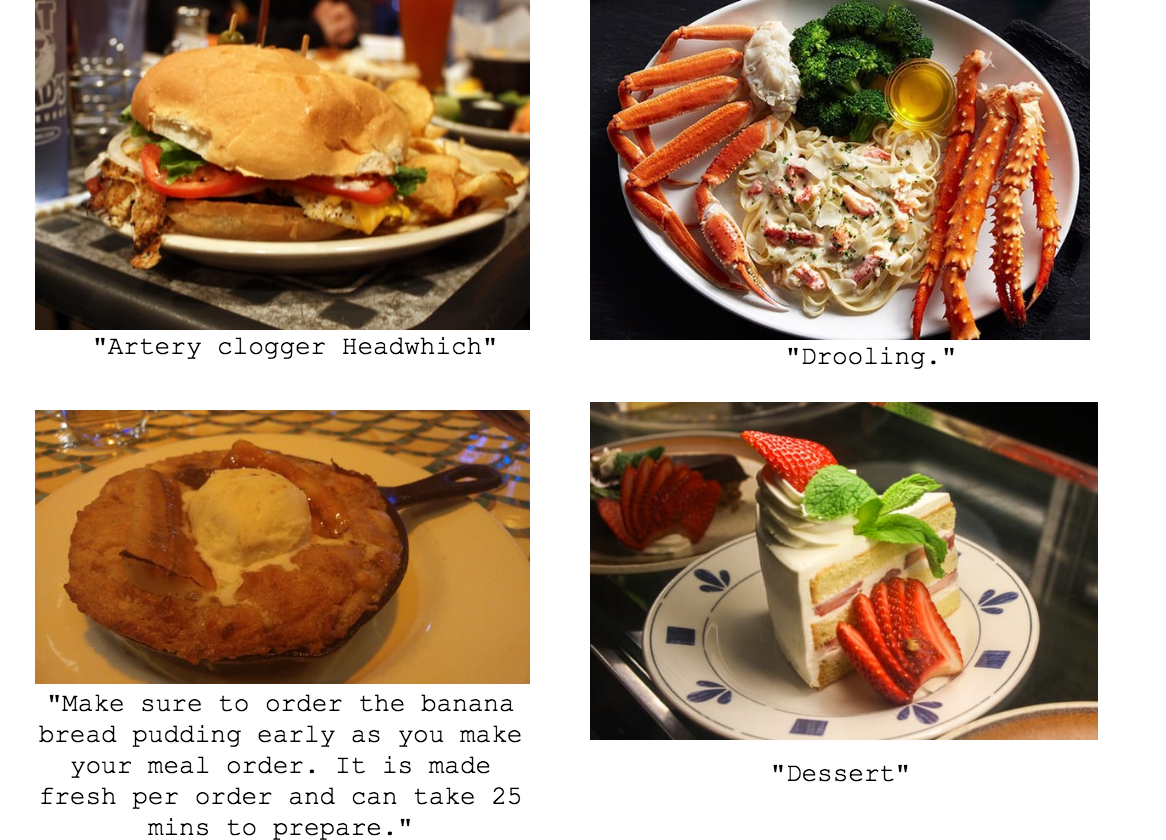}
\captionsetup{singlelinecheck=off}
\caption[list=off]{
Sample images from the Yelp image dataset were chosen to exemplify the varying noise sources in the dataset.
\begin{itemize}
\item \emph{Top-left:} Rarely encountered noun phrases such as `Artery clogger' and apocryphal words like `Headwhich' make identifying words tied to an image's content difficult.
\item \emph{Top-right:} Captions unrelated to the image content.
\item \emph{Bottom-left:} Wordy captions are likely to have content that is particular to an image and does not generalize to images with similar content.
\item \emph{Bottom-right:} Over simplified captions. This training example may cause a neural network to learn a link between the features of strawberry shortcake and the word \emph{dessert}, which describes a much larger class of dishes.
\end{itemize}
}
\end{figure}

\section{METHODOLOGY}

This section describes the preprocessing and feature extraction steps we took to prepare the Yelp dataset for training. We then outline the different model architectures we tested and a custom scheme for generating captions from unseen images using these models. 

\subsection{Extracting Image Embeddings}

A NIC system must incorporate features extracted from both an input image and a partial caption \footnote{Initially, the `partial caption' is a special token that signifies the beginning of the predicted sequence, \texttt{<startseq>}. Then, one predicts the next word of the sequence and feeds it back to the network until the token \texttt{<endseq>} is predicted by the network, which marks the end of the predicted sequence.} in order to predict the next word of the caption. This requires that one compress an input image into a fixed-length vector, which encodes the visual features of the image - thus forming a sort of \emph{image embedding.}

One approach to doing so is to construct a model architecture in which input images pass through many layers - responsible for extracting informative features from the images - before incorporating these image features with the rest of the NIC model - and then training the entire model using backpropagation.  

The problem with this approach is that powerful neural networks that extract complex features from images often have millions of parameters. Thus, a NIC trained to extract image features would be slow to train and have a propensity to overfit.

\begin{figure*}[ht]
\centering
\includegraphics[width=1\linewidth]{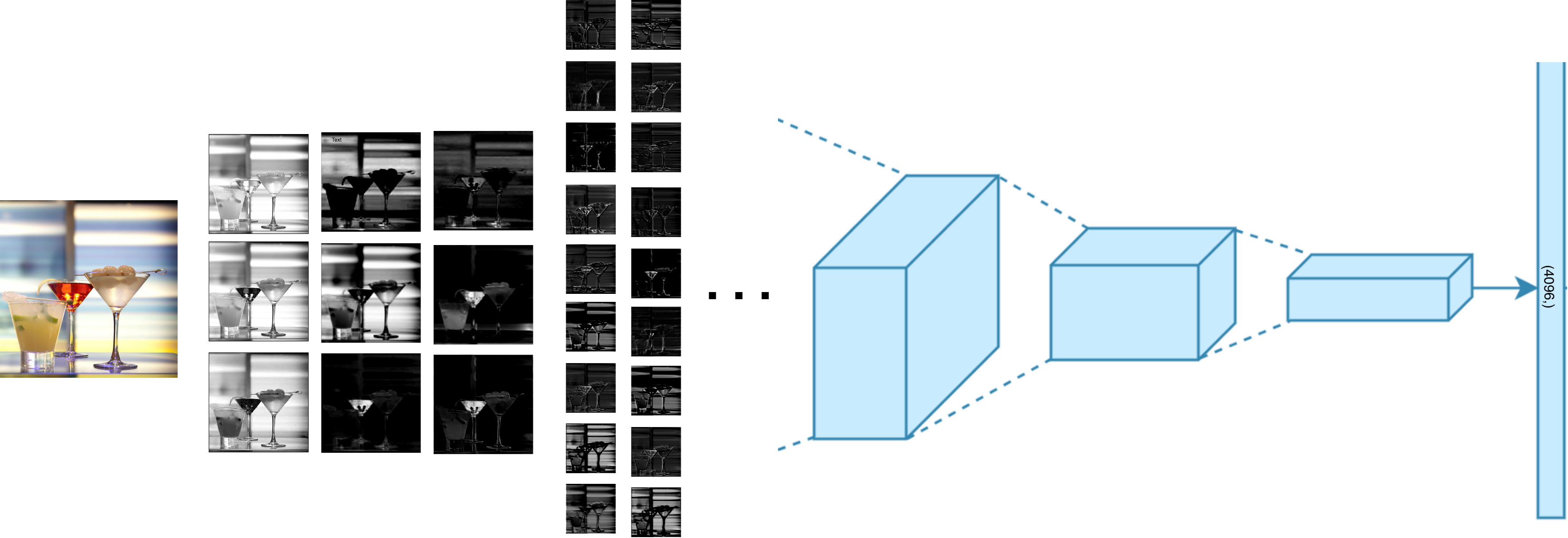}
\caption{Vector representations of images, or ``image embeddings", are created by saving an intermediate activation of the VGG16 network, which can detect rich perceptual features, such as edges, blur and hue. Combined with linguistic features, these embeddings are used to generate input image captions. Thus, the abridged VGG16 network may be considered the \emph{encoder} in the \emph{Encoder-Decoder} framework.}
\label{fig:test1}
\end{figure*}

Instead, to create image embeddings for each of the training images, we utilized the pre-trained weights of the VGG16 network (Simonyan \& Zisserman; 2014), available in the \code{keras} deep learning library (Chollet et al.; 2015). This network, with 138,357,544 parameters and trained on over 1.3 million images from the ImageNet dataset (Russakovsky et al.; 2015), secured the first and the second places in the ImageNet ILSVRC-2014 localization and classification competitions, respectively. Thus, it is reasonable to assume that this network can extract complex features from diverse images. 

The last 3 layers of the VGG16 architecture are fully connected, with dimensions $4096$, $4096$, and $1000$, respectively. To generate an image embedding for each image in the Yelp dataset, we first removed the last two fully connected layers from VGG16 and used the predictions of this abridged model as the image embeddings [figure 5]. We dropped the last two layers because while the first layers of the network are responsible for feature extraction, the final layers are specialized for the task they were trained for - namely, ImageNet classification. Thus, omitting the final two layers allows me to encode each image as a $4,096$ dimensional vector, which is informative and generalizes to the image captioning task.

\subsection{Caption Text Preprocessing}

A widespread challenge in Natural Language Processing (NLP) tasks is that corpus vocabularies are often sparse - meaning that few words occur many times, any many words occur few times\footnote{Empirical studies repeatedly show that the frequency of a word in a corpus is inversely proportional to its frequency rank -  a result commonly referred to as \emph{Zipf's Law} (Wyllys, R; 1981)}. Sparse vocabularies lead to models with many parameters and make generalizing patterns learned during training harder. 

Thus, we performed a cursory cleanse of the captions provided in the Yelp dataset with the goal of reducing vocabulary size. This cleanse is not thorough and is likely a bottleneck in our final captioning system. 

The steps we took - in order - are:
\begin{itemize}
\item Lowercase all characters.
\item Remove newline characters.
\item Replace the character \texttt{\&} with the word \texttt{and}. This is so the system will identify phrases like \texttt{sweet \& spicy} and \texttt{sweet and spicy} to be identical.
\item Convert strings which match a simple regex pattern of a website domain\footnote{The regex pattern used to detect website domains is \texttt{[a-z\textbackslash:\textbackslash/\textbackslash.0-9]+\textbackslash.(org|com|net)}, using the python \texttt{re} library.} to the token \texttt{website}. This is because two images in the Yelp dataset are unlikely to refer to the same website domain, leading to tokens of low frequency and a larger vocabulary.
\item For the same reasons, we replaced all tokens with numeric characters with the token \texttt{num}.
\item Using the python \code{unidecode} library, we re-encoded all characters to ASCII characters. This way, phrases with/without accents like \textit{gruyère soufflé} and \textit{gruyere souffle} are consolidated to the same tokens. The \code{unidecode} library also implements functions for transliterating tokens from foreign languages to English characters, further reducing vocabulary size.
\item Remove all remaining punctuation.
\item Prepend all captions with the token \texttt{<startseq>}, and postpend with the token \texttt{<endseq>}, to mark the beginning and end of each caption.
\end{itemize}
After these steps, the processed captions contain 30,012 unique terms. 

\subsection{Formatting Data for Training}

Section 3.A discusses how one can conceptually fit the image captioning task into the probabilistic framework. Equation (2) shows that under mild assumptions, the probability of a caption being correct, given an image, can be decomposed into the probability of each token given the previous tokens and the image: 

\begin{equation*}
\log P(S_1, ... S_k | I; \theta) = \sum_{t = 1}^k{\log P(S_t|S_1 ... S_{t-1}, I; \theta)}
\end{equation*}

Typically, NIC systems model $P(S_t|S_1 ... S_{t-1}, I; \theta)$ with a recurrent neural network (RNN) sub-architecture. That is, the RNN is responsible for predicting the next token, given the previous tokens in the caption and the image; this way, the RNN plays the role of a language model. 

Inspired by (2), we reformat the training captions to be compatible with a language modeling task. First, we extract an image embedding for each image in the dataset (section 5.A), and preprocess the text of the corresponding caption (section 5.B). Then, for each embedding/caption pair, we create a new training example for each token in the caption, where each token plays the role of the response, and the  covariates\footnote{\textit{Covariates}, in this context, refers to the features which are trained upon. Also referred to as \textit{independent variables} or \textit{predictors}, and typically denoted as $\mathbb{X}$ in the statistics literature.} are the tokens that occur previously in the original caption, as well as a copy of the corresponding image embedding [figure 6].

\begin{figure}[H]
\centering
\includegraphics[width=1\linewidth]{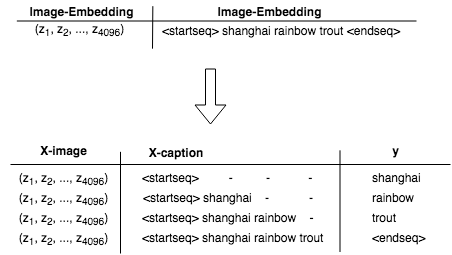}
\caption{An example of how an image-embedding/caption pair is reformatted to suit a language modeling task. Each token (except for the first \texttt{<startseq>} token is treated as a response, where the covariates are the previous tokens in the sequence and a copy of the image embedding.}
\label{fig:test1}
\end{figure}

Captions that exceed 15 tokens in length are truncated to 15 tokens. Captions that are shorter than 15 tokens are padded by a special \texttt{null} token. This way, all captions have a uniform length of 15 tokens.

After performing this expansion, the original 100,807 images with captions become 675,289 image-embedding/sub-caption pairs. When splitting the dataset into training/validation splits (section 6), we split the data \textit{before} performing the expansion detailed above. This is to ensure that the models are trained on complete captions.

\subsection{Model Architectures}

\begin{figure*}[h]
\centering
\begin{minipage}{.5\textwidth}
  \centering
  \includegraphics[width=.8\linewidth]{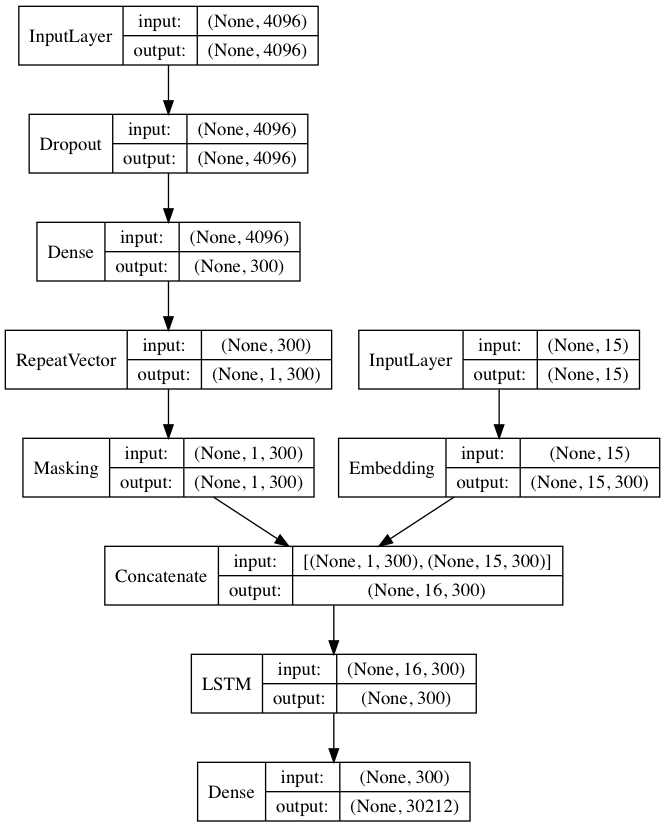}
  \label{fig:test1}
\end{minipage}%
\begin{minipage}{.5\textwidth}
  \centering
  \includegraphics[width=.8\linewidth]{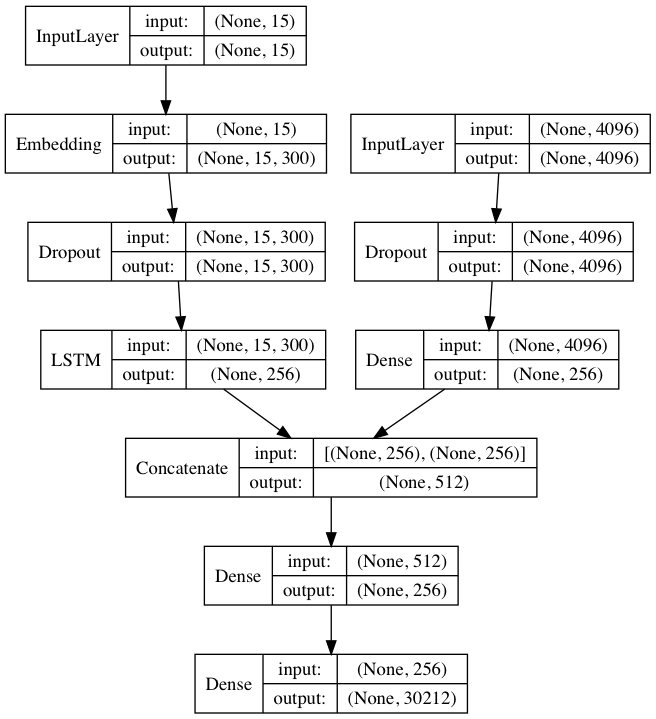}
  \label{fig:test2}
\end{minipage}
\caption{Inject model (left) and merge-concat models (right). In the injected model, image features are treated as the first ``word" in the caption and then passed to the LSTM cell. In the merge-concat model, image and caption features are processed separately, concatenated, and then passed to a dense layer to form predictions. \\
The merge-add layer (not shown above) differs from the merge-concat model only in that image and caption features are \textit{added} instead of \textit{concatenated}. \\
Layer names above refer to \href{https://keras.io/layers/about-keras-layers/}{\texttt{keras} layer classes}.}
\end{figure*}

In this project, we experiment with three model architectures - two \textit{merge models} and one \textit{inject model}, using the terms defined in section 3.C. See Figure 7 for architectural details.

For the injected model, image embeddings are fed to a dense\footnote{Dense layers are fully connected layers.} layer of output dimension 300. The activation of this layer is then prepended to the word embeddings from the caption input to form a matrix of dimension $(16,300)$, which is then fed to an LSTM layer with a hidden state of 300 dimensions. In other words, the image features are treated as the first ``word" in the sequence before being processed by the RNN language model. The output of this LSTM is fed to a dense layer with a softmax activation to predict the next word in the sequence. We refer to this architecture as simply the ``inject model."

After being transformed into word embeddings in both merge models, the captions are passed to an LSTM cell with a hidden state of 256 dimensions. Image embeddings are passed to a dense layer of output dimension 256. 

The activation of this dense layer is combined with the output of the final LSTM cell, though the way they are combined differs between the two merge models. In the first, the two activations are \textit{concatenated}, yielding a single 512-dimensional vector fed to a dense layer to form predictions. This model is called the \textit{``merge-concat"} model. In the second, the two activations are \textit{added}, yielding a vector of dimension 256, passed to a dense layer to form predictions. This model is called the \textit{``merge-add"} model.

The justification for experimenting with both the merge-concat and merge-add model is that the final dense layer of the merge-concat model has many parameters, as a 512-dimensional layer is fully connected to the final 30012-dimensional layer. The merge-add model halves the number of parameters in the final weight matrix, as a 256-dimensional layer is fully connected to the final 30012-dimensional layer.

\subsection{Inference as Beam Search}

The models detailed in section 5.D each take in two vectors as input - an image generated by passing an image through the VGG16 network and the integer representation of a (partial) caption. The output of the models is a 30012-dimensional vector - where each element represents the predicted probability of a certain word appearing next in the caption. 

Given this setup, it is not obvious how to generate a new caption for an image previously unseen by the network, as partial captions are not available during the inference stage. 

A naive method for generating new captions is \textbf{greedy selection}. Starting with an image feature and a caption with only the start token \texttt{<startseq>}, one repeatedly uses a trained model to predict the probability of each word being next in the caption. Then, the word with the highest predicted probability is added to the caption. This process is repeated until the maximum sequence length is reached\footnote{In our models, we used a maximum sequence length of 15 tokens.}, or the special token \texttt{<endseq} is selected, marking the end of a sequence [algorithm 1].

\begin{algorithm*}[h]
\caption{Inference: Greedy Selection}\label{greedy-sampling}
\begin{algorithmic}[1]
\Procedure{GreedySelect(img-features, model)}{}
\State $\textit{caption} \gets \text{[}<startseq>\text{]}$
\Comment Initialize caption as start token.
\While {Length(caption) $<$ 15}
\Comment Repeat until generated caption is of maximum length.
\State $\textit{predictions} \gets \texttt{model.predict(img-features, caption)}$
\Comment Vector of predicted probabilities
\State $\textit{next-word} \gets \text{argmax(\textit{predictions})}$
\Comment Predicted next word is the one with the highest predicted probability.
\If {\textit{next-word} == \texttt{<endseq>}}
\State \textbf{break}
\Comment If end token is predicted, return caption as-is.
\Else
\State $\textit{cation} \gets \textit{cation}\text{.append(\textit{next-word})}$
\EndIf
\EndWhile
\Return \textit{caption}
\EndProcedure
\end{algorithmic}
\end{algorithm*}

\begin{algorithm*}[h]
\caption{Inference: Beam Search}\label{beam-search}
\begin{algorithmic}[1]
\Procedure{BeamSearch(img-features, model, $\beta$,$\kappa$, $\alpha$)}{}
\State $\textit{population} \gets \text{[}<startseq>\text{]}$
\Comment Initialize population as a single `starter' caption.
\State $i \gets 0$
\Comment Iteration Number
\For{$ i < 15 $}
\For{\text{each candidate caption } $S \in$ \textit{population}}
\If{Last token in candidate == \texttt{<endseq>}}
\State \textbf{break}
\EndIf
\State $\textit{predictions} \gets \texttt{model.predict(img-features, caption)}$
\Comment Vector of predicted probabilities
\State Add top $\kappa$ predicted words to caption to create $\kappa$ new candidates
\EndFor
\State Truncate population top $\beta$ candidates, according to quality metric.
\State $i \gets i + 1$
\EndFor
\Return \textit{population}
\EndProcedure
\end{algorithmic}
\end{algorithm*}

In each iteration, greedy selection selects the most probable word and adds it to the generated caption. It does not allow for adding words that may be ``suboptimal" at a given iteration but then enables the model to make a prediction it is very confident of in a later iteration. Since our goal is to generate a sequence of tokens $\textbf{S}$ which well approximates the \textit{joint} probability of each of the tokens, $\textbf{S}^* = \argmax_{\textbf{S}}{P(S_1, S_2, ..., S_k|I)}$, it seems that this scheme is too rigid. 

A more flexible method of inference is \textbf{Beam Search}. Instead of building up a single caption as greedy selection does, one always maintains $\beta$ candidate solutions, where $\beta$ is called the \textit{beam width.} The set of candidate solutions is often called the \textit{population} of candidates. At each iteration, each candidate solution in the population is expanded into $\kappa$ candidates by adding the words with the top $\kappa$ predicted probabilities to the candidate, where $\kappa$ is called the \textit{neighborhood size.} After each candidate in the population is expanded into $\kappa$ new candidates, they are sorted according to some quality metric, and the top $\beta$ candidates are moved on to the next iteration. 

In this work, we define a custom quality metric inspired by the personal model of what a ``good" caption looks like. A good prediction should be descriptive and relevant to an image's content. It should also be succinct and avoid adding predicted tokens if the model is not ``confident" in these predictions. 

The quality metric is calculated as follows: for each candidate caption $S \equiv (S_1, S_2, ... S_k)$, keep track of the probabilities of each token $\Omega_S = (\omega_1, \omega_2, ... \omega_k)$\footnote{In this notation, $\omega_t$ is the predicted probability of the token $S_t$.}, predicted by the model during earlier iterations of the beam search. Then, the quality score of the sequence $S$ is a weighted sum of the predicted probabilities $\Omega_S$, where the weight multiplied by $\omega_t$ is $\alpha^t$, for some $\alpha \leq 1$: 

\begin{equation}
\text{Score}\big ( (\omega_1, \omega_2, ... \omega_k) \big ) = \sum_{t = 1}^k \omega_t\cdot\alpha^t
\end{equation}

This measure rewards candidates whose tokens are predicted with high probability but discounts later tokens due to the geometric decrease in weights, $\alpha^t$. Thus, $\alpha$ is a sort of regularization hyperparameter - small values of $\alpha$ lead to terse captions, while large values of $\alpha$ lead to captions at or near the maximum token length [figure 8].  

\begin{figure}[h]
\centering
\includegraphics[width=.6\linewidth]{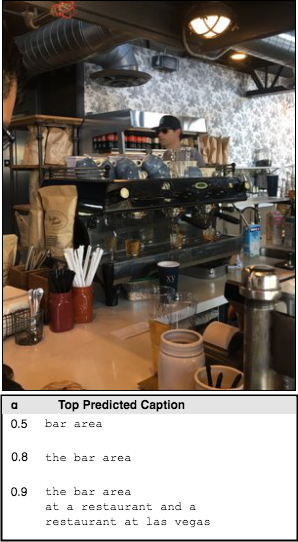}
\caption{Top caption predictions using beam search and the inject model for different values of $\alpha$. Small values of $\alpha$ lead to overly terse captions, while large values of $\alpha$ lead to wordy and repetitive captions.}
\label{fig:test1}
\end{figure}

G greedy selection is a special case of beam search, where $\beta = \kappa = \alpha = 1$.  

\section{EXPERIMENTS}

Before training, we randomly set aside 20\% of the 100,807 images with captions to form a validation set; the remaining 80\% form the training set. We kept this split consistent in all experiments to gauge the relative effectiveness of the different models and hyperparameter configurations. 

Although many hyperparameters should be tuned in both the inject and merge models\footnote{See section 8 for a discussion of hyperparameters to tune in future work}, we only experimented with different combinations of model architectures and $\alpha$ - discussed in section 5.E. For each inject, merge-concat, and merge-add models, we generated and evaluated predictions using beam search with $\alpha \in \{.6, .7, .8\}$, leading to 9 total fits. 

\subsection{Training details}

We trained all three model architectures using stochastic gradient descent with Nesterov momentum, with the learning rate $\eta = .01$, and the momentum term $\mu = .9$, and a learning rate decay rate\footnote{At iteration $t$, the learning rate $\eta$ is updated: $\eta \gets \frac{\eta}{1  + t\cdot10^{-6}}$.} of $10^{-6}$. We did not tune these hyperparameters, as we observed that the training loss converged steadily with this configuration [figure 9]. We used categorical-cross-entropy as the objective function.

\begin{figure}[h]
\centering
\includegraphics[width=.75\linewidth]{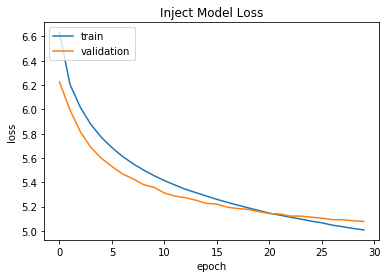}

\includegraphics[width=.75\linewidth]{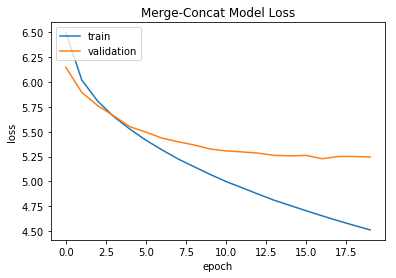}

\includegraphics[width=.75\linewidth]{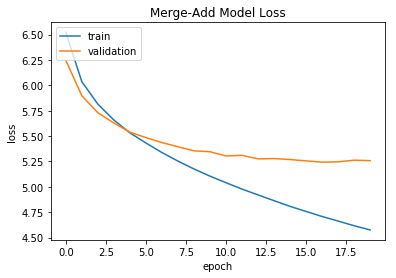}

\caption{Training and validation losses for the inject, merge-concat, and merge-add models.}
\label{fig:test1}
\end{figure}

We trained each model for 30 epochs. We used an early stopping scheme - stopping training when validation loss does not decrease for two consecutive epochs. Each training epoch took around 8 minutes on an NVIDIA Kepler GK104 GPU hosted on Amazon's AWS EC2 service. 

The merge-concat and merge-add models converged faster than the injected model; after 18 epochs, the early stopping scheme suspended training. On the other hand, training loss decreased much more slowly for the injected model, and training continued for the full 30 epochs. The inject model achieved the lowest validation loss at the end of training - evidence that although it is slower, it does a better job of generalizing to new examples.

\section{Results}

In this section, we explore the efficacy of each of the nine NIC system configurations we discuss in section VI. we discuss two metrics for quantitatively evaluating the performance of a NIC system - \textit{BLEU} and \textit{ROUGE}, and compute these metrics on each configuration's predictions. We then perform a brief qualitative analysis - discussing some unexpected behaviors of our system and their potential causes.

\subsection{Quantitative Analysis}

\begin{table*}[h]
\centering
\label{my-label}
\caption{Validation set prediction quality scores.}
\begin{threeparttable}
\begin{tabular}{|
>{\columncolor[HTML]{C6D6D7}}c |
>{\columncolor[HTML]{E4EFEF}}c |c|c|c|c|c|c|c|}
\hline
\cellcolor[HTML]{CBCBE0}\textbf{Model}                          & \cellcolor[HTML]{CBCBE0}\textbf{$\alpha$} & \cellcolor[HTML]{CBCBE0}\textbf{BLEU-1} & \cellcolor[HTML]{CBCBE0}\textbf{BLEU-2} & \cellcolor[HTML]{CBCBE0}\textbf{BLEU-3} & \cellcolor[HTML]{CBCBE0}\textbf{BLEU-4} & \cellcolor[HTML]{CBCBE0}\textbf{ROUGE-L} & \cellcolor[HTML]{CBCBE0}\textbf{Terms Generated} & \cellcolor[HTML]{CBCBE0}\textbf{Average Caption Token Length } \\ \hline
\cellcolor[HTML]{C6D6D7}                                        & \textbf{.6}                               & .0687                                   & {\ul .0295}                             & {\ul .0137}                             & {\ul .0095}                             & .0885                                    & 501                                              & 3.17                                                    \\ \cline{3-9} 
\cellcolor[HTML]{C6D6D7}                                        & \textbf{.7}                               & .0693                                   & .029                                    & .0129                                   & .0083                                   & .0962                                    & 559                                              & 4.31                                                    \\ \cline{3-9} 
\multirow{-3}{*}{\cellcolor[HTML]{C6D6D7}\textbf{Inject}}       & \textbf{.8}                               & {\ul .0711}                             & .0283                                   & .0121                                   & .0082                                   & {\ul .1172}                              & {\ul 644}                                        & 7.57                                                    \\ \hline
\cellcolor[HTML]{C6D6D7}                                        & \textbf{.6}                               & \textbf{.084}                           & \textbf{.0384}                          & \textbf{.0207}                          & \textbf{.0128}                          & \textit{.1133}                           & \textit{1191}                                    & 6.41                                                    \\ \cline{3-9} 
\cellcolor[HTML]{C6D6D7}                                        & \textbf{.7}                               & \textit{.0832}                          & .0367                                   & \textit{.0198}                          & \textit{.0125}                          & \textit{.1229}                           & \textit{1294}                                    & 8.1                                                     \\ \cline{3-9} 
\multirow{-3}{*}{\cellcolor[HTML]{C6D6D7}\textbf{Merge-Concat}} & \textbf{.8}                               & \textit{.0781}                          & \textit{.033}                           & \textit{.018}                           & \textit{.0122}                          & \textbf{.139}                            & \textbf{1406}                                    & 10.49                                                   \\ \hline
\cellcolor[HTML]{C6D6D7}                                        & \textbf{.6}                               & {\ul .0812}                             & .0373                                   & {\ul .0192}                             & {\ul .0125}                             & .1096                                    & 961                                              & 4.96                                                    \\ \cline{3-9} 
\cellcolor[HTML]{C6D6D7}                                        & \textbf{.7}                               & .0782                                   & {\ul \textit{.0338}}                    & .0171                                   & .011                                    & .1197                                    & 1058                                             & 7.31                                                    \\ \cline{3-9} 
\multirow{-3}{*}{\cellcolor[HTML]{C6D6D7}\textbf{Merge-Add}}    & \textbf{.8}                               & .0735                                   & .0304                                   & .0152                                   & .001                                    & {\ul .1311}                              & {\ul 1147}                                       & 9.98                                                    \\ \hline
\end{tabular}
\begin{tablenotes}
\item BLEU-N, ROUGE-L and number of terms generated for each model configuration. Boldface represents the highest score achieved amongst any model configuration for that metric. Underscore represents the best score achieved for all models with the same architecture (inject, merge-concat or merge-add), and italics represent the best score achieved amongst configurations with the same value of $\alpha$. 
\end{tablenotes}
\end{threeparttable}
\end{table*}

\begin{figure*}[h]
\centering
\includegraphics[width=.45\linewidth]{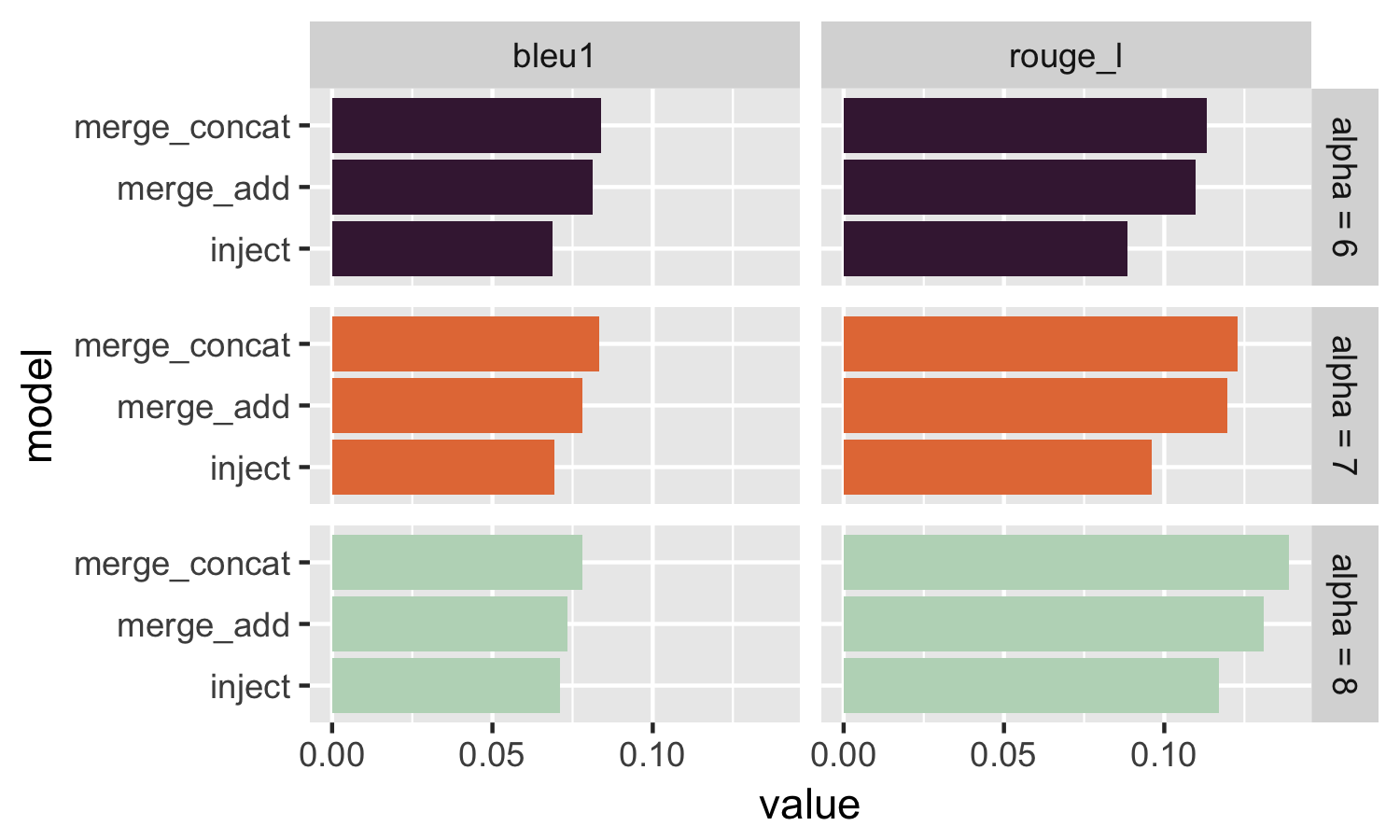}
\includegraphics[width=.45\linewidth]{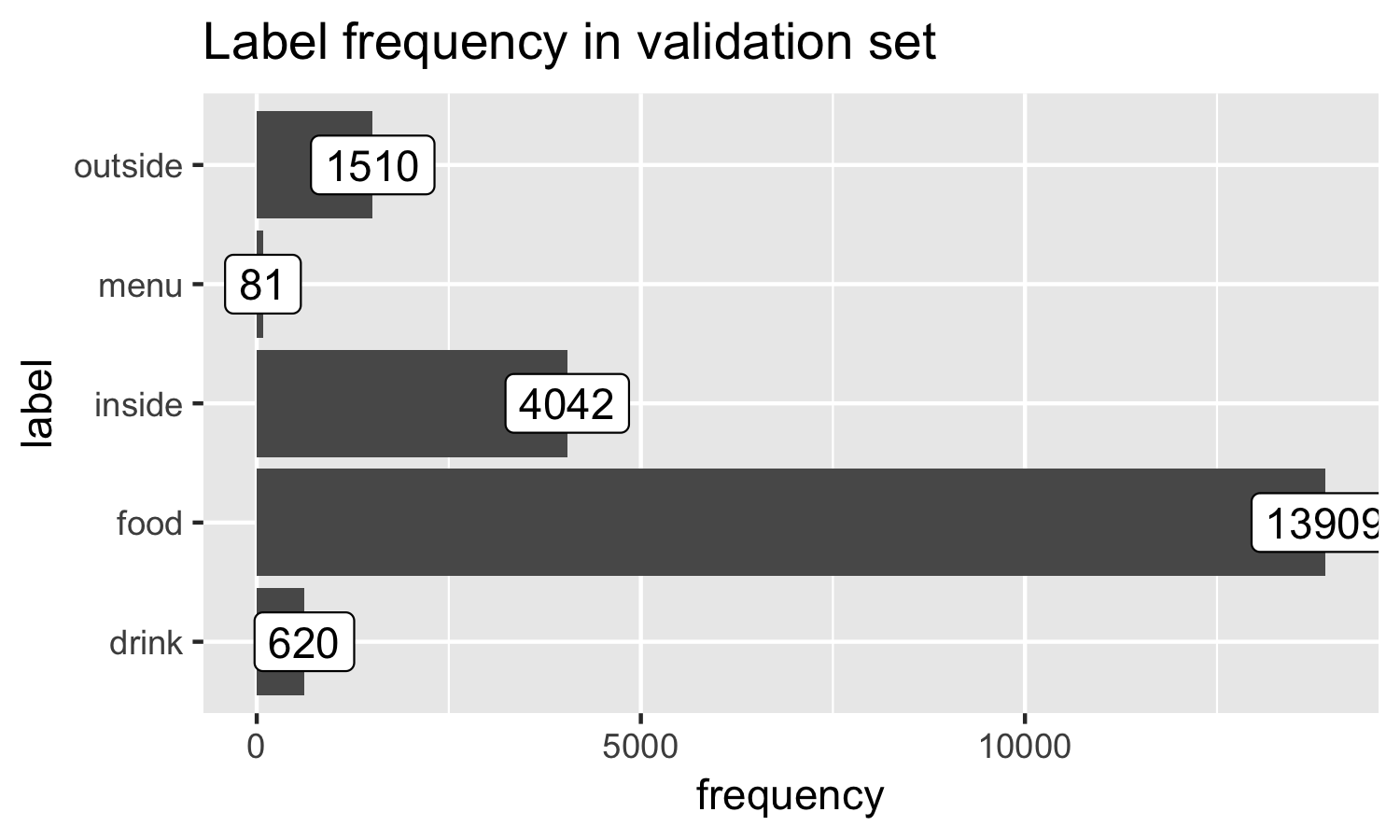}

\includegraphics[width=.32\linewidth]{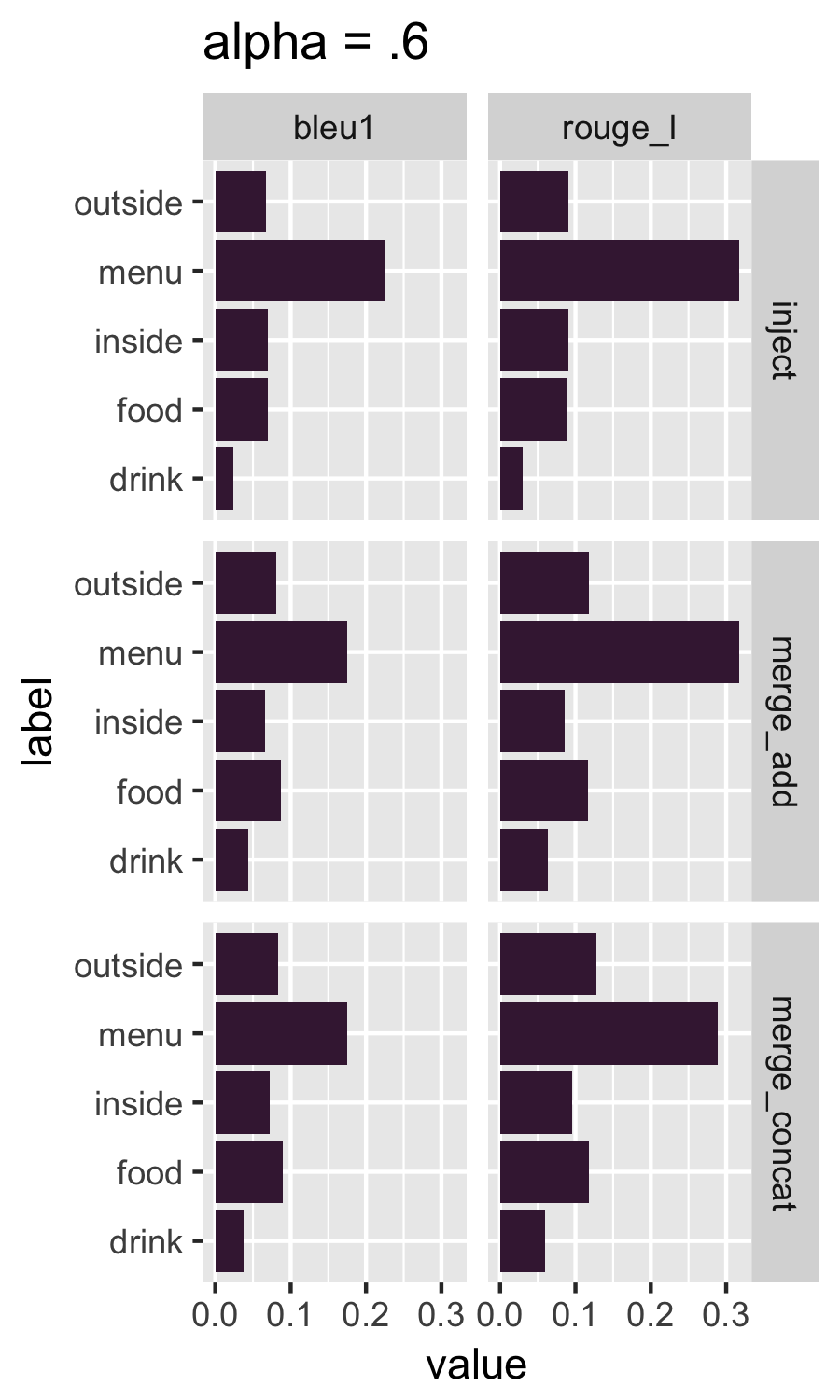}
\includegraphics[width=.32\linewidth]{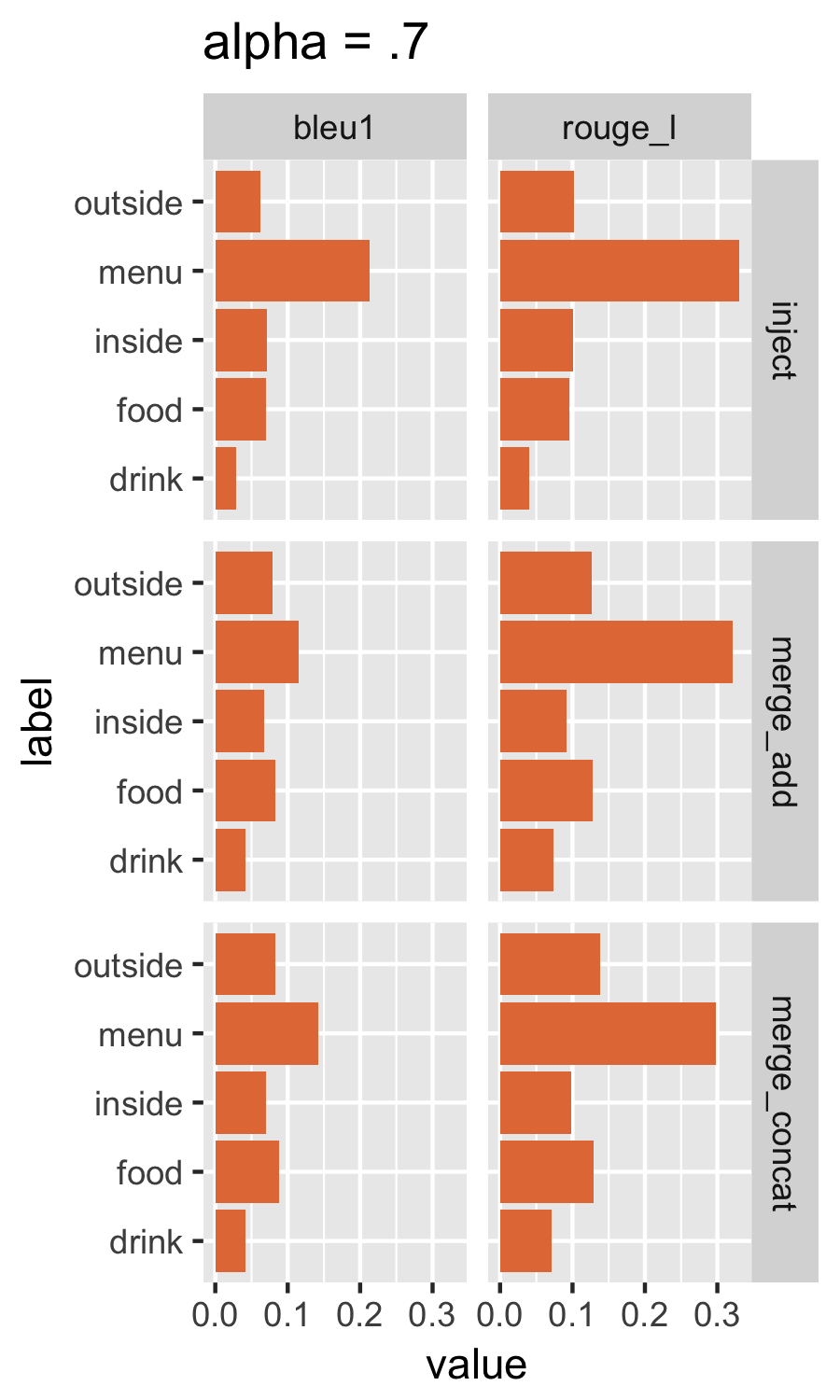}
\includegraphics[width=.32\linewidth]{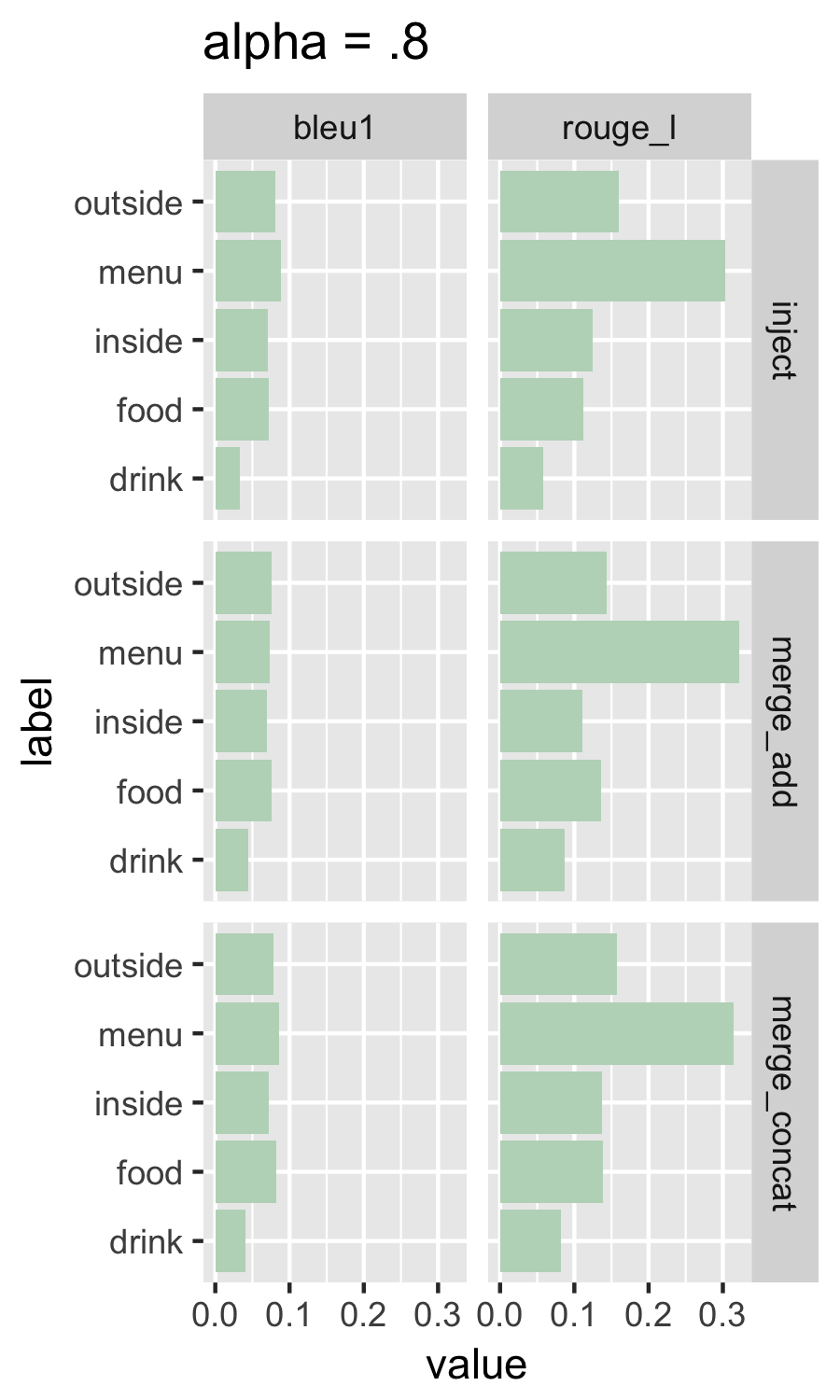}

\caption{\textbf{Top-left:} BLEU-1 and ROUGE-L scores grouped by $\alpha$, to highlight differences between model architectures. \textbf{Top-right:} Label frequency in the validation set. \textbf{Bottom:} BLEU-1 and ROUGE-L achieved by each model on validation examples grouped by label, split by the value of $\alpha$.}
\end{figure*}

Defining paradigms for evaluating the quality of generated captions remains a challenge in active research (for example, Vedantam et al., 2014). Perhaps the most widely adopted metric for evaluating machine translation and image captioning system predictions is \textit{BLEU} (Papineni et al., 2002). BLEU-N counts n-grams in the predicted text and the ``true" reference text. Since it is a precision-based metric, short predictions are naturally favored by the most basic version of BLEU. Thus, most implementations (including the one we used\footnote{We used the BLEU implementation \href{https://www.nltk.org/_modules/nltk/translate/bleu_score.html}{from python's \texttt{nltk} library.}}) penalize short predictions, and predictions that use only frequently occurring terms. 

BLEU is agnostic to n-gram ordering and thus cannot measure the coherence of a predicted caption. Although it is not perfect, the score is widely adopted partly because it is inexpensive to compute, and it has been shown to correlate highly with human evaluation. We compute BLEU-1, BLEU-2, BLEU-3, and BLEU-4 for each model configuration as a proxy for measuring how relevant the predictions are to the images' content.

A robust NIC system should not only form predictions that are related to the image content; it should also generate diverse captions utilize a substantial vocabulary, and describe multiple objects/relationships in images. Thus, a metric that complements the precision-based BLEU is the recall-based \textit{ROUGE} (Lin; 2004). We report ROUGE-L as a proxy of the diversity of the captions generated by each model configuration. We also report the number of unique terms generated by each configuration as a measure of the vocabulary size utilized by each model [table 1].

Large values of $\alpha$ yield high ROUGE scores, and low values of $\alpha$ yield high BLEU scores, holding the model architecture fixed. This is as expected, as high values of $\alpha$ yield more wordy predictions, which leads to higher recall but lower precision. 

Interestingly, the merge-concat and merge-add models yield a much larger vocabulary than the injected model and higher ROUGE scores. In fact, for any given $\alpha$, the merge-concat yields a vocabulary that is more than double the size of the injected model. Notice, however, that the average length of captions generated by the merge models is much larger than those of the injected model. Thus, the fact that the merge models yield superior ROUGE scores and larger vocabulary sizes may suggest that these models indeed tend to utilize richer vocabularies. It may also suggest that merge models are more prone to generating ``wordy" captions, and perhaps more strict regularization (smaller $\alpha$) would lead to predictions more similar to those generated by the injected models.

The best BLEU scores are also achieved by the merge models, which suggests that these models are more likely to predict words that are present in the ground truth.

It's also interesting to note the differences in BLEU and ROUGE scores achieved by the models on validation examples from different categories [figure 10]. For $\alpha \in \{.6, .7\}$, all three model architectures achieved substantially higher BLEU-1 and ROUGE-L scores for images in the \textit{menu} category, though this does not hold for $\alpha = .8$. This result is unreliable, however, as only 81 images (0.4\%) of the images in the validation set are labeled \textit{menu.} Of the remaining 4 categories that are better populated, the models achieve the smallest BLEU and ROUGE scores in for example in the \textit{drink} section, which may suggest that images with this label are harder to predict accurately.

\subsection{Qualitative Analysis}

\begin{figure*}
\begin{center}
\textbf{Everything looks like Chicken and Waffles... Or does it?}\par\medskip
\begin{tabular}{lll}
 \includegraphics[scale = .23]{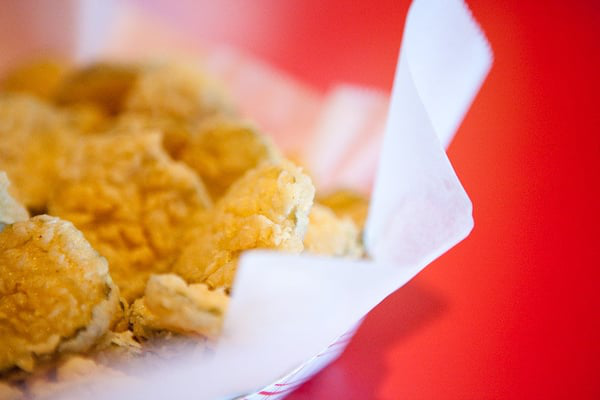}& \includegraphics[scale = .23]{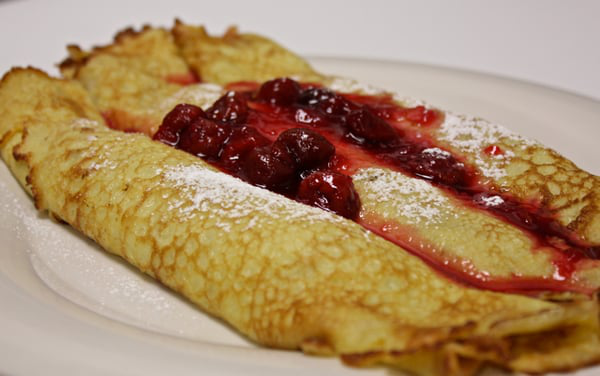} &  \includegraphics[scale = .23]{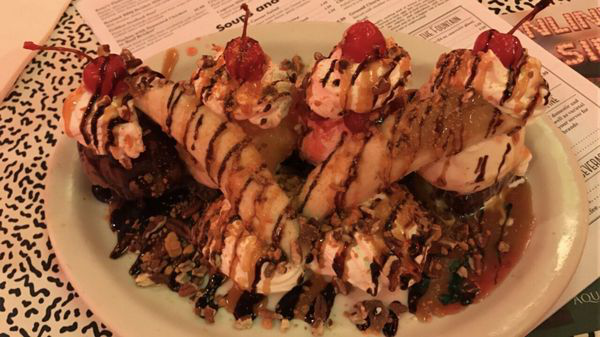}\\
\includegraphics[scale = .23]{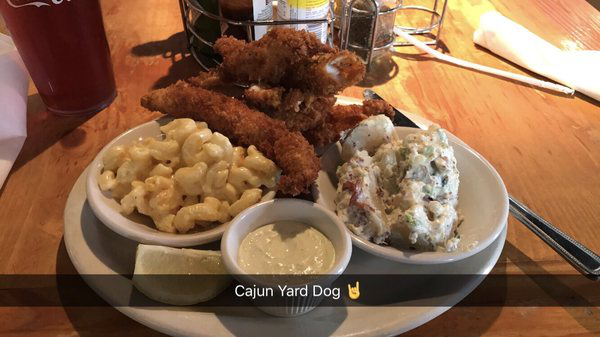} & \includegraphics[scale = .23]{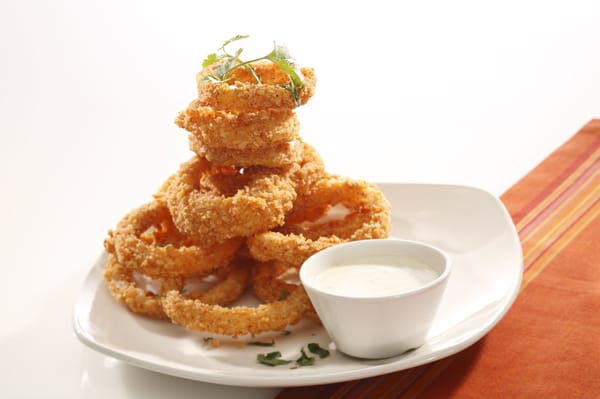} & \includegraphics[scale = .23]{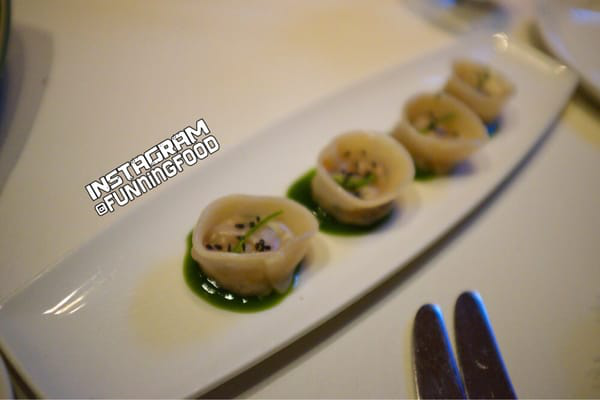} \\
 \includegraphics[scale = .23]{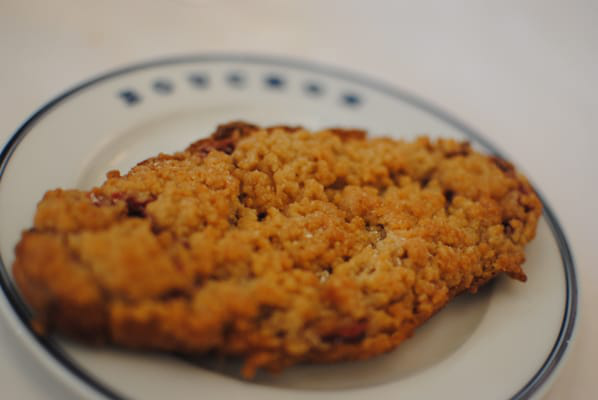}& \includegraphics[scale = .23]{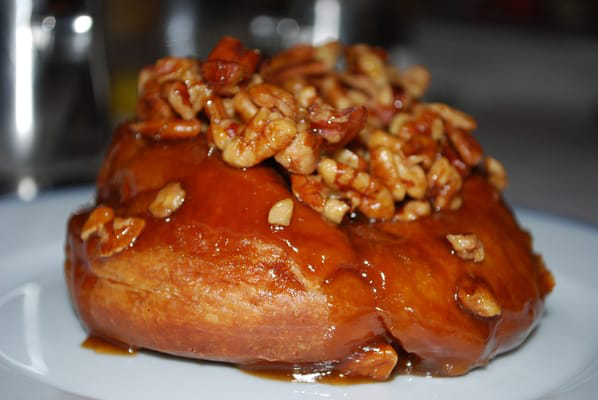} & \includegraphics[scale = .23]{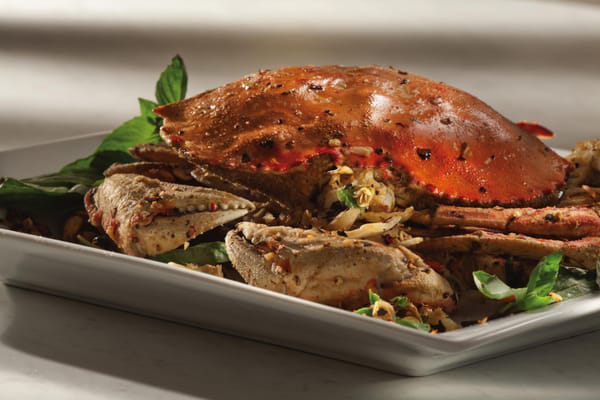}
\end{tabular}
\end{center}
\caption{Sample images the final system predicts are ``chicken and waffles." The system predicts around one in five images - often or images that look nothing like the American dish. This behavior resembles a policy of \textit{``when in doubt: chicken and waffles.''}}
\end{figure*}

Although the study of BLEU and ROUGE scores [table 1, figure 10] would lead one to believe that the merge models perform better than the injected models, manually reviewing these models' predictions reveals that the injected model performs much better in practice. 

The merge models' predictions are often long and nonsensical and repeat common phrases like \textit{``special offer, limited time only"} and \textit{``with a side of mac n cheese."} Our theory for why this occurs is that the merge models are severely overfitting, as evidenced by the training trajectories\footnote{For the merge models, the training loss is much lower than the validation loss - a clear sign of overfitting. This phenomenon is not as severe for the injected model.} in Figure 9. Thus, subtle features in an image may cause a merge model to regurgitate a sequence of words present in the training data with high confidence, even though those words do not relate to the image. 

Using the inject architecture, the system generated the most relevant and appropriately lengthed predictions, with $\alpha = .6$. This is the configuration used in the final system.

Although the inject architecture is more well-behaved than the merge models, it too exhibits bizarre behavior - most notably its eagerness to label images as \textbf{chicken and waffles.} Of the 20,162 validation images, the final system predicts that 3,866 of them (19.2\%) contain the string ``chicken and waffles" [figure 11].

This is not a result of the images' content but rather of the sparseness of the training captions and the method with which we train the network. 

Consider a naive agent that completely ignores the image at hand when making a prediction and behaves as an n-gram-based language model. Starting with no prior image nor caption information, the most reasonable\footnote{If we assume the first word of a caption does not depend on subsequent words, this is the maximum likelihood estimate.} prediction for the first word in a sequence is the word that leads the training captions most frequently - \texttt{the}, followed by the second most common word to lead the training captions - \texttt{chicken} [figure 12]. If one were to use this agent to generate captions using beam search with a beam width $\beta \geq 2$ (as we have), then the population after the first iteration would contain the partial caption \texttt{[chicken]}.

\begin{figure}[H]
\centering
\includegraphics[width=.8\linewidth]{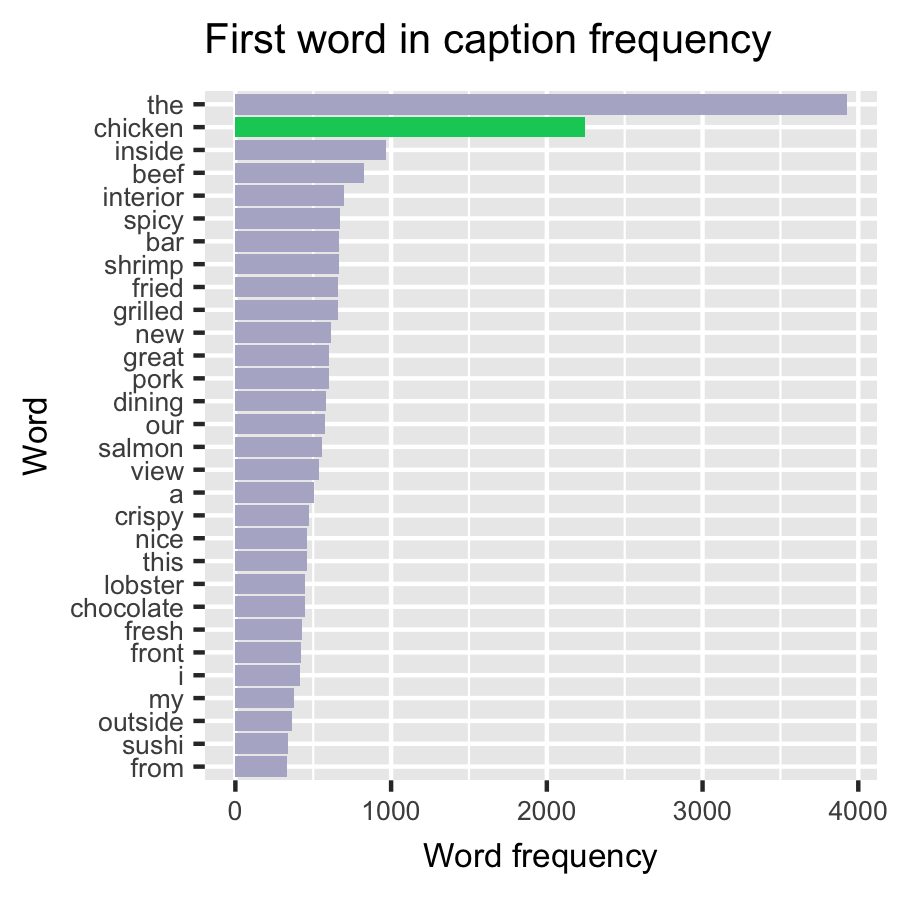}
\caption{Words that lead training captions most frequenlty.}
\label{fig:test1}
\end{figure}

To select the second word to add to this caption, the agent would find the word that is preceded by the word \texttt{chicken} most frequently in the training captions, which is \texttt{and} [figure 13]. Similarly, the next iteration of the beam search would lead to the word \texttt{waffles} being added to the caption, as it is the second most frequently word to follow the bigram \texttt{chicken and} [figure 14]. Thus, the naive agent generates the string \texttt{chicken and waffles}. 

\begin{figure}[H]
\centering
\includegraphics[width=.8\linewidth]{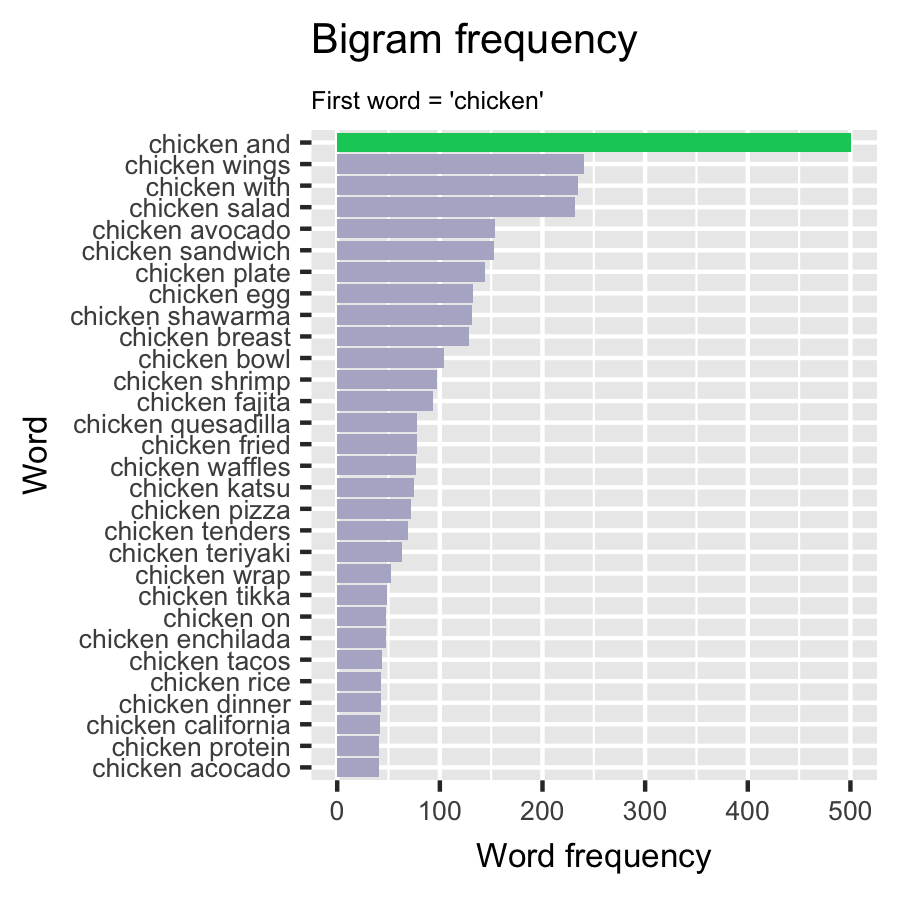}
\caption{Most common bigrams - given that the first word is \texttt{"chicken"}.}
\label{fig:test1}
\end{figure}
\begin{figure}[H]
\centering
\includegraphics[width=.8\linewidth]{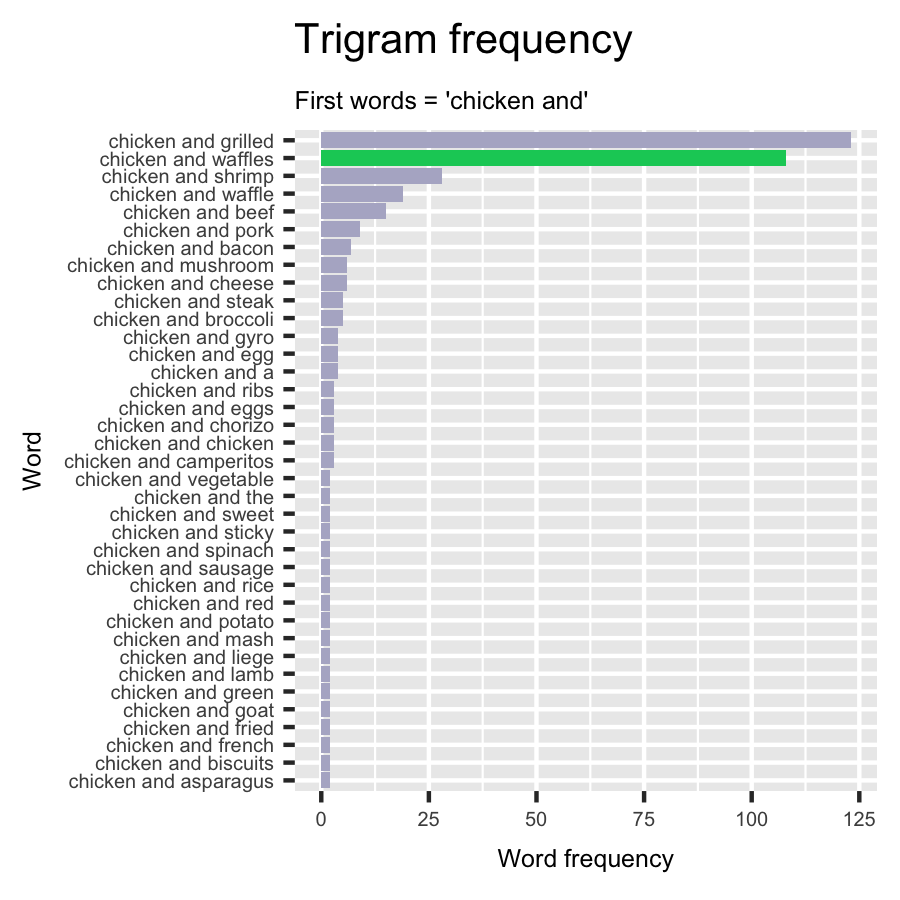}
\caption{Most common trigrams - given that the first two words are \texttt{"chicken and"}.}
\label{fig:test1}
\end{figure}

\newpage

Thus, an intuitive explanation for why our NIC system emits the string \textit{"chicken and waffles"} so frequently isn't that it sees chicken and waffles in the image. Instead, \textit{it has no idea what is in the image}, and so it predicts the string \textit{"chicken and waffles"} because it behaves as a naive n-gram-based language model. 

This problem is exacerbated by the sparseness of the token frequency table the training captions generate. Token frequencies follow a power rule with a very long tail, which motivates the language model component of an NIC to predict the few frequently occurring n-grams, regardless of an image's content. 

\section{CONCLUSION}

In this technical report, we've discussed the inner workings of a simple neural image captioning system. After a short time playing with the application, it becomes clear that the system can detect objects in an input image and form a sequence of natural language conditioned on the image content.

However, the system we've built is far from perfect, mainly due to the lack of due diligence in our methodology. In our experiments, we compare three NIC model architectures and conclude that the ``inject" architecture (section 5.D) generates the most relevant and succinct predictions. However, this result is not legitimate, as we did not thoroughly tune the hyperparameters associated with each model. In future work, some of the important hyperparameters we would like to tune are:
\begin{itemize}
\item Hyperparameters associated with the optimization scheme, such as the learning rate $\eta$, batch size, momentum ter $\mu$, etc. 
\item Model architecture and the complexity of each model, for example, by tuning the number of hidden layers, the output dimension of each dense layer, the hidden state size of the LSTM layer, etc. This is especially critical for the merge models, which display clear evidence of overfitting [figure 9].
\item Different transfer learning methods - for example, by experimenting with networks other than VGG16 for encoding images and using word embedding models other than Word2Vec. It may be useful to allow the network to modify pre-trained word embeddings or learn word embeddings from scratch.
\item Use of regularization to control overfitting.
\item Ignoring terms that have a corpus frequency lower than a given threshold as a means of reducing the vocabulary size of the training captions.
\end{itemize}

Besides tuning these hyperparameters, a more important and difficult task we would like to undertake in future work is to process the Yelp dataset to be more consistent and ``well-defined." Currently, the captions of image captions are irregular; a caption may describe the content of an image, a user's experience at an establishment, or advertise a restaurant's impressive menu and social media presence. One cannot expect a system to behave consistently after it is trained on inconsistent data, hence the phrase \textit{``garbage in, garbage out."} Refining the Yelp dataset for the purpose of building an image captioning system, for example, by categorizing images by the semantic content of their captions, is an undertaking that warrants its own separate - and likely very beneficial - research project. 

Due to the rise of user-friendly deep learning frameworks, public access to pre-trained neural network models and word embeddings, and inexpensive cloud computing services, anyone can build applications that would have been considered science fiction a decade ago in a few hundred lines of code and a modest time/price budget. This democratization of artificial intelligence will undoubtedly bring great value to companies like Yelp, embrace the public's creativity, and enable them to build useful systems by sharing their data. We hope our work in this paper will inspire novice machine learning practitioners to take on unconventional machine learning tasks that are not yet well-understood, such as neural image captioning. Furthermore, we hope this work will motivate internet companies to share their data and invite the public to share their work and express their creativity. In this new age of democratized artificial intelligence, companies that take on this transparent and open mindset will enjoy substantial value generated from the masses of passionate learners.

\bibliographystyle{IEEEtran}
\bibliography{main}
\end{document}